\documentclass[sigconf]{acmart}
\usepackage[T1]{fontenc}
\usepackage{CJKutf8}

\AtBeginDocument{%
  \providecommand\BibTeX{{%
    \normalfont B\kern-0.5em{\scshape i\kern-0.25em b}\kern-0.8em\TeX}}}

\copyrightyear{2023}
\acmYear{2023}
\setcopyright{acmlicensed}\acmConference[CIKM '23]{Proceedings of the 32nd ACM International Conference on Information and Knowledge Management}{October 21--25, 2023}{Birmingham, United Kingdom}
\acmBooktitle{Proceedings of the 32nd ACM International Conference on Information and Knowledge Management (CIKM '23), October 21--25, 2023, Birmingham, United Kingdom}
\acmPrice{15.00}
\acmDOI{10.1145/3583780.3615151}
\acmISBN{979-8-4007-0124-5/23/10}

\begin{document}

\title{Generating News-Centric Crossword Puzzles As A Constraint Satisfaction and Optimization Problem}

\author{Kaito Majima}
\authornote{Both authors contributed equally to this research.}
\email{kaito.majima@nex.nikkei.com}
\orcid{0009-0001-0834-1200}
\affiliation{
  \institution{Nikkei Inc.}
  \city{Otemachi}
  \state{Tokyo}
  \country{Japan}
}

\author{Shotaro Ishihara}
\email{shotaro.ishihara@nex.nikkei.com}
\authornotemark[1]
\orcid{0009-0001-0366-6807}
\affiliation{
  \institution{Nikkei Inc.}
  \city{Otemachi}
  \state{Tokyo}
  \country{Japan}
}

\renewcommand{\shortauthors}{Majima and Ishihara}

\begin{abstract}
    Crossword puzzles have traditionally served not only as entertainment but also as an educational tool that can be used to acquire vocabulary and language proficiency.
    One strategy to enhance the educational purpose is personalization, such as including more words on a particular topic.
    This paper focuses on the case of encouraging people's interest in news and proposes a framework for automatically generating news-centric crossword puzzles.
    We designed possible scenarios and built a prototype as a constraint satisfaction and optimization problem, that is, containing as many news-derived words as possible.
    Our experiments reported the generation probabilities and time required under several conditions.
    The results showed that news-centric crossword puzzles can be generated even with few news-derived words.
    We summarize the current issues and future research directions through a qualitative evaluation of the prototype.
    This is the first proposal that a formulation of a constraint satisfaction and optimization problem can be beneficial as an educational application.
\end{abstract}



\begin{teaserfigure}
  \includegraphics[width=17cm]{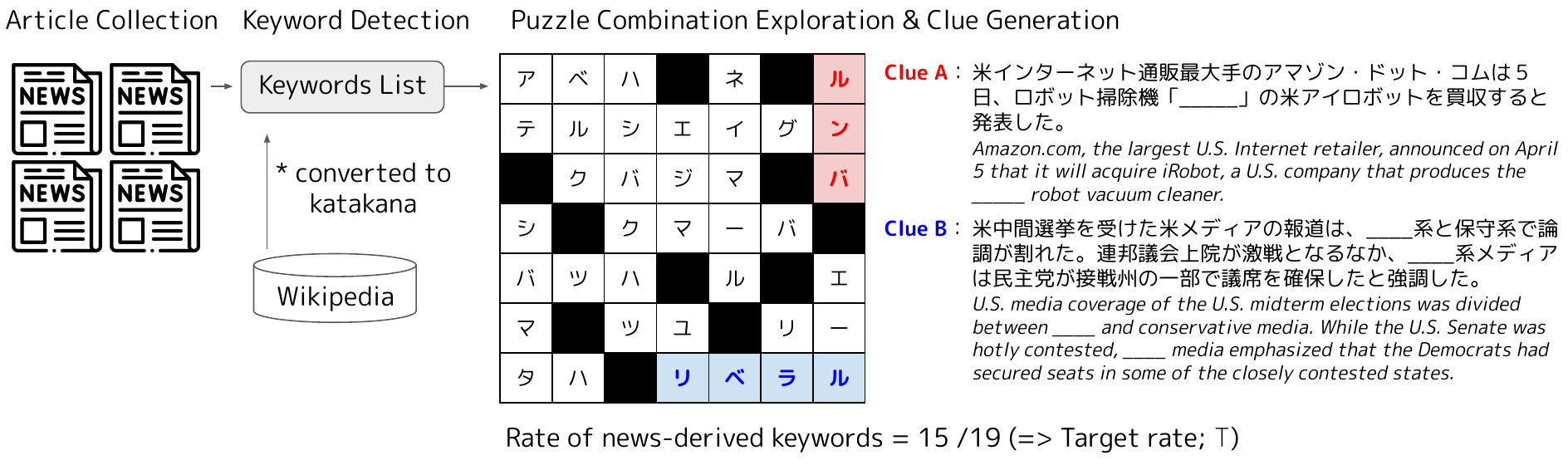}
  \centering
  \begin{CJK}{UTF8}{ipxm}
  \caption{
    Overview of the proposed framework.
    From the collected articles and external resources, nouns are detected and masked by named entity recognition to create answers and clue pairs.
    Finally, puzzle combinations are generated to obtain a crossword puzzle.
    We present only two clues with English translations.
    The answers are ルンバ (\textit{Roomba}) and リベラル (\textit{liberal}).
  }
  \label{fig:overview}
  \end{CJK}
\end{teaserfigure}

\maketitle

\section{Introduction}

Crossword puzzles have been popular as entertainment and have contributed to education.
The literature has reported contributions such as acquiring critical thinking~\cite{Ritonga2021-wq, Khaira2021-xe} and vocabulary~\cite{Franklin2003-cl, Suryawati2018-cv}.
On the other hand, the manual creation of crossword puzzles is a complex task that requires exceptional skills. Therefore, there is a growing interest in automatic question generation aimed at cost reduction and ensuring continuity~\cite{Chali2015-tm, Kurdi2020-rz, Heilman2010-ie,De_Kegel2020-jl}.
Automatic generation is important, especially when considering personalization to enhance educational effectiveness.
For example, crossword puzzles that include a large number of news-derived words are expected to have a stimulating effect on people's interest in the news.
However, the framework for creating personalized crossword puzzles for educational applications has not been sufficiently researched.

This study proposes a strategy to automatically generate personalized crossword puzzles that contain many words on particular \emph{topics}.
Focusing on news as a \emph{topic}, we design a framework that can automatically generate crossword puzzles that contain as many news-derived words as possible (Figure \ref{fig:overview}).
This procedure can be implemented using a combination of existing natural language processing and mathematical optimization methods.
However, the combination of techniques is not obvious and requires sophisticated design, experimentation, and evaluation~\cite{Wang2022-qa}.
We are the first to identify that a specific problem design in mathematical optimization, namely \emph{a constraint satisfaction and optimization problem}~\cite{Brailsford1999-nw}, is beneficial for this application.

In summary, this research makes the following contributions:
\begin{enumerate}
  \item We designed possible scenarios to increase interest in the news and prototyped the automatic generation of news-centric crossword puzzles as a constraint satisfaction and optimization problem (Section \ref{sec:proposed}).
  \item Our experiments showed that news-centric crossword puzzles can be generated even with a small number of news-derived words (Section \ref{sec:Experiments}).
  \item We described our findings from a qualitative evaluation and outlined current issues and future directions (Section \ref{sec:Discussion}).
\end{enumerate}

\section{Related Work}
\label{sec:RelatedWork}

This section reviews related studies from three perspectives.

\subsection{Puzzle Combination Exploration}
\label{subsec:RelatedCrossword}

There is a long history of research on automatic crossword puzzle generation~\cite{De_Kegel2020-jl}.
Generating crossword puzzles is known to be an NP-hard problem~\cite{Bonomo2015-ou,Botea2021-rb,Bulitko2021-jb}.
Mazlack et al. proposed a letter-by-letter fulfillment approach in 1976~\cite{Mazlack1976-wz} and Ginsberg et al. employed a word-by-word filling strategy in 1990~\cite{Ginsberg1990-em}.
Meehan et al. compared these two strategies and concluded that the word-by-word approach is more effective~\cite{meehan1997constructing}.
Bulitko et al. grouped the research surrounding crossword puzzles into three categories~\cite{Bulitko2021-jb}: solving a puzzle~\cite{Littman2002-sm}, generating without a score~\cite{Ginsberg1990-em,Botea2007-bn}, and generating for a higher score~\cite{Bonomo2015-ou,Botea2021-rb,Bulitko2021-jb,Audemard2020-yj}.
In the third category, Douglas et al. used genetic algorithms to generate crossword puzzles~\cite{Bonomo2015-ou}, and most others adopted the mathematical optimization approach.

Our method of puzzle combination exploration follows prior work~\cite{Botea2021-rb} in the third category and solves a constraint satisfaction and optimization problem.
Specifically, the task is to generate crossword puzzles with a score containing many news-derived words and to fill in the answers word-by-word.
We emphasize that this is the first to propose this problem formulation as an educational application aimed at increasing words derived from specific topics, in this case, news.
Unlike some previous studies~\cite{Bulitko2021-jb,Dakowski2022-if}, the placement of black cells is not explored.

\subsection{Crossword Puzzles for Education}

There is limited but gradually growing attention to automatic question generation in the context of education, with the goal of reducing costs and providing continuity.
Esteche et al. proposed a method to extract words and their definitions from Spanish news and automatically create crosswords~\cite{Esteche2016-db}.
They also utilize external dictionaries and fill the gaps readily based on the score values attached to the words.
Some studies have examined the educational benefits of crossword puzzles~\cite{Franklin2003-cl, Ritonga2021-wq, Khaira2021-xe,Suryawati2018-cv}.

Our study is similar to the work of Esteche et al.~\cite{Esteche2016-db} in that we combined multiple technologies to generate crossword puzzles from news articles for educational purposes.
Recent automatic question generation systems increasingly use neural networks for end-to-end learning~\cite{Sun2018-pm,Zhang2019-nz}, but these systems face challenges in controlling question difficulty and considering different sources of knowledge.
Wang et al. conducted a needs assessment survey of 11 faculty members at seven universities and argued that we need a combination of fine-grained techniques and careful design to implement question generation in practice~\cite{Wang2022-qa}.

\subsection{News for Education}

News content has traditionally been associated with education and is considered effective for improving reading comprehension and interest in current affairs.
Newspapers have been widely used as educational tools, for example, there is an international program called \textit{Newspapers in Education}~\cite{DeRoche1981-um,Claes2009-uk}.
As the issue  of information overload grows and the problem of fake news has become more apparent, it has become increasingly important to develop the skill of information literacy among the general public.

However, interest in the news has been rapidly declining, particularly among young people\footnote{\url{https://reutersinstitute.politics.ox.ac.uk/digital-news-report/2022}}.
While crossword puzzles are popular content in news media\footnote{\url{https://www.nytco.com/press/both-cooking-and-games-reach-1-million-subscriptions/}}, there is a limited research focus on their educational effectiveness~\cite{Ferrer-Conill2020-ic}.
This study provides a new role of crossword puzzles within the news media.

\section{Proposed Framework}
\label{sec:proposed}

This section proposes a strategy for automatically generating personalized crossword puzzles that contain many words on particular topics.
In particular, we describe a scenario for the news as a topic and a prototyping methodology.

\subsection{Possible Scenarios}
\label{subsec:Scenario}

This study assumes a scenario where crossword puzzles can motivate people to read news articles and thereby increase their interest in the news.
Examples include: 1) encouraging people to keep up with current events based on information published regularly, such as news content found in morning and evening newspapers, and 2) reviewing the comprehension of current events based on news articles read by each individual.
These examples can be customized.
For example, there are options to focus on the genre (e.g., economics, politics), and limit the types of words (e.g., company, person).

\subsection{Prototyping}
\label{subsec:Prototype}

We describe a prototype for automatic crossword puzzle generation that can be applied to specified scenarios.
The procedure consists of four steps, as shown in Figure \ref{fig:overview}: article collection, keyword detection, clue generation, and puzzle combination exploration.

\subsubsection{Article Collection}

The first step is the gathering of articles according to the specified scenario and objective.
For example, to encourage people to grasp current events, it is a good idea to gather information published regularly in morning and evening newspapers.
If individual browsing histories are available, it is also possible to check the level of understanding of current events by gathering news articles read by each individual.
We also have to prepare external resources because the news articles alone do not provide a sufficient volume of words to generate crossword puzzles.
We use dumped data from Wikipedia\footnote{\url{https://meta.wikimedia.org/wiki/Data_dumps}}.

\subsubsection{Keyword Detection}

The second step is the detection of keywords to be used as answers.
Candidate keywords are generated by extracting nouns from the text body using Named Entity Recognition (NER)~\cite{Yadav2018-gb}.
The same process can be used to extract keywords from external resources.
The prototype procedure proposed in this study is almost language-independent.
However, if there are multiple character types, we should convert them into a specified type.
For example, Japanese text contains several character types including Chinese characters, hiragana, katakana, and alphabetical letters.

\subsubsection{Clue Generation}

The third step is the generation of clues corresponding to the answers.
We mask the body text (fill-in-the-blank) as a simplified method.
For the case of clue B in Figure~\ref{fig:overview}, when \textit{liberal} is extracted, we can create the clue sentence as follows: “U.S. media coverage of the U.S. midterm elections was divided between \texttt{[Answer]} and conservative media.”

\subsubsection{Puzzle Combination Exploration}

Finally, we explore puzzle combinations from the set of answers, by following prior work~\cite{Botea2021-rb} to solve a problem formulated as the constraint satisfaction and optimization problem.
Specific constraints are as follows: 
\begin{itemize}
  \item The placement of black cells (which do not contain letters) and slots (which contain letters) are given and fixed.
  \item All slots must be filled with letters without any conflicts.
  \item The solution must contain at least \texttt{T} \% of the news-derived words.
\end{itemize}
We search for optimal choices efficiently and terminate the process when a solution is found.
The algorithm proceeds to fill in the answers one by one within the constraints, using backtracking, a depth-first search approach~\cite{Tarjan1971-hx}.
The controllable hyperparameters are the crossword puzzle size (vertical; \texttt{V} and horizontal; \texttt{H}), the placement of the black cells (\texttt{P}), the target rate of the news-derived words (\texttt{T}), and time limit ($\bar{t}$).
The reason for setting \texttt{T} is that a crossword puzzle with more news-derived words is appropriate.
Note that we terminate the search every 10 seconds and retry with different random states for improving the success probability.

\section{Experiments}
\label{sec:Experiments}

This section shows experimentally that the proposed method can generate personalized crossword puzzles from real-world data sets under certain conditions.
The purpose of the experiments is to automatically generate crossword puzzles in a variety of different settings and to examine the following:
\begin{itemize}
  \item How does the target rate (\texttt{T}) affect the success probability and the generation time? (Figure~\ref{fig:experimental_results} \& \ref{fig:discussion})
  \item How does the placement of black cells (\texttt{P}), specifically the number of black cells, affect the generation time? (Figure~\ref{fig:dist_generation_time})
\end{itemize}

\subsection{Data set}

In this study, we used a real-world data set, \textit{Nikkei}\footnote{\url{https://aws.amazon.com/marketplace/seller-profile?id=c8d5bf8a-8f54-4b64-af39-dbc4aca94384}}, a Japanese financial news corpus.
There are several additional challenges in handling Japanese data sets compared to English: converting Chinese characters to katakana, more than twice as many letter choices as the English alphabet, etc.
Most previous studies have used English, so using Japanese is a valuable case study.

\subsection{Experiment Setup \& Implementation}

Here we describe the setup and implementation of an experiment to create the outputs.
In the experiment, we fix the size of the crossword puzzle (\texttt{H} and \texttt{V}) at 7, which is one of the standard sizes in Japanese crossword puzzles\footnote{\url{https://vdata.nikkei.com/nikkeithestyle/crossword/}}.
Then we explore the placement of the black cells (\texttt{P}), and the target rate of news-derived words (\texttt{T}).

The prototype consists of four steps, and the steps except for puzzle combination exploration are common in our setup.
We prepared 2,006 news-derived words and 449,895 keywords from Wikipedia articles through the steps of article collection and keyword detection.
We used \textit{WikiExtractor}\footnote{\url{https://github.com/attardi/wikiextractor}} for pre-processing Wikipedia text.
News-derived words are proper nouns extracted by NER.
We considered the headwords as keywords in the external resource Wikipedia articles.
All keywords are converted to katakana by \textit{pykakasi}\footnote{\url{https://codeberg.org/miurahr/pykakasi}}.
In the clue generation step, we created fill-in-the-blank questions using each keyword and the body text.
The clues in Figure~\ref{fig:overview} were generated by the fill-in-the-blank method from news articles\footnote{\url{https://www.nikkei.com/article/DGKKZO63247650W2A800C2EA2000}}\footnote{\url{https://www.nikkei.com/article/DGKKZO65863840Q2A111C2FFJ000}}.

During the puzzle combination exploration step, we recorded the success probabilities and generation times for each pattern:
\begin{itemize}
  \item The target rate of the news-derived words (\texttt{T}): Every 10 \% from 10 to 100 \%. We started with a small \texttt{T} and ended our search when we could no longer succeed in generating additional combinations.
  \item The placement of the black cells (\texttt{P}): Random generation of 10 patterns of 9-12 black cells each; 40 patterns in total.
\end{itemize}
Success is defined as generating a singular crossword puzzle that meets the given conditions within a time constraint of 300 seconds.

\subsection{Results of Puzzle Combination Exploration}

\begin{figure}[t]
  \centering
  \includegraphics[width=7.5cm]{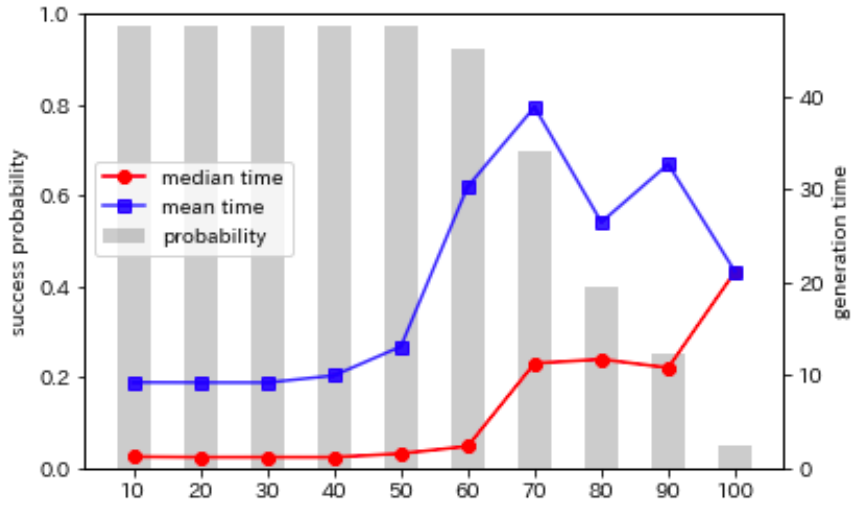}
  \caption{
    Success probability and generation time (in seconds) by target rate (\texttt{T}).
    The higher the rate, the lower the success probability and the longer the generation time.
  }
  \label{fig:experimental_results}
\end{figure}

\begin{figure}[t]
  \centering
  \includegraphics[width=6.6cm]{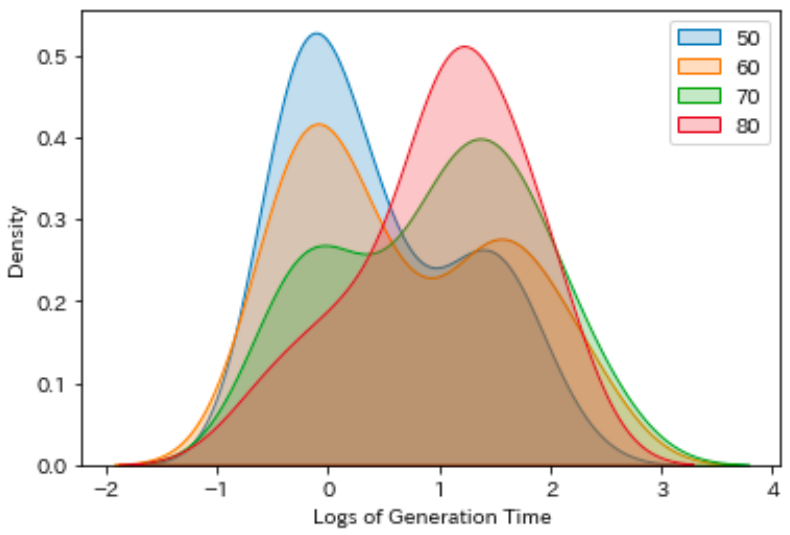}
  \caption{
    Distribution of generation times when the target rate (\texttt{T}) is higher.
    When \texttt{T} is above 50, the distribution of generation times becomes unstable.
  }
  \label{fig:discussion}
\end{figure}

\begin{figure}[t]
  \centering
  \includegraphics[width=6.6cm]{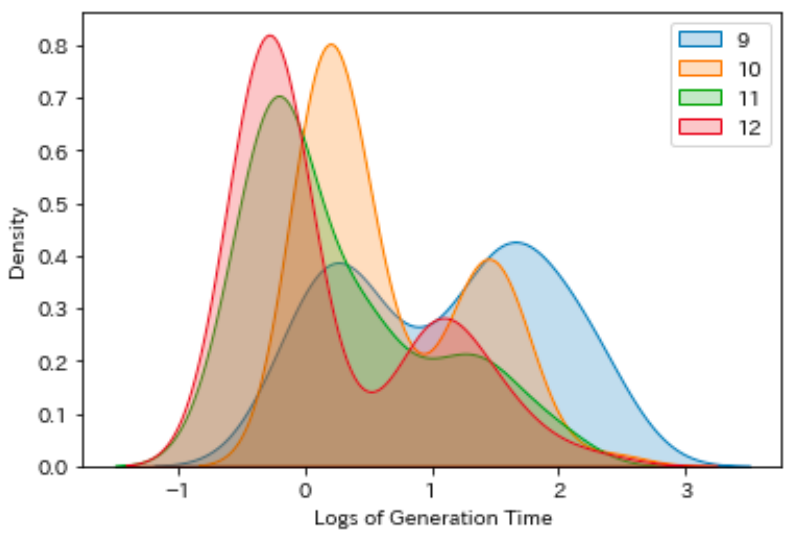}
  \caption{
    Distribution of generation times by the number of black cells.
    The more black cells there are, the shorter the generation time.
  }
  \label{fig:dist_generation_time}
\end{figure}

Figure \ref{fig:experimental_results} shows success probability and generation time by target rate (\texttt{T}).
When \texttt{T} was less than 60 \%, crossword puzzles could be generated with a probability of around 90 \%.
In that case, the median generation time is less than 10 seconds, which is quite fast.
As \texttt{T} constraints increase, success probability decreases, and generation time increases.
Figure \ref{fig:discussion} shows the distribution of generation times when \texttt{T} is higher.
We can see that the generation time increases unstably when \texttt{T} is above 50.
Figure \ref{fig:dist_generation_time} shows the distribution of generation times by the number of black cells.
The more black cells there are, the smaller the search space becomes, and thus the shorter the search time.

Such puzzle combination exploration is helpful in setting appropriate hyperparameters for practical use.
For example, in this case, setting T to 50 and ensuring a minimum of 11 black cells is expected to result in a success probability of over 90 \% and generation times typically within 10 seconds.
Considering the gap between the number of keywords from news articles (2,006) and Wikipedia (449,895), it is a contribution of mathematical optimization that the success probability of crossword puzzles with \texttt{T=50} was over 90 \%.

\section{Future Research Directions}
\label{sec:Discussion}

We demonstrated the prototype to approximately 20 people of several demographic groups and asked for their reactions.
The target group included researchers in natural language processing and crossword puzzle creators.
This section lists future research directions based on the findings from this qualitative evaluation.

\paragraph{Improved Article Collection}

Our study selected news as one of the topics, but there are other extensions.
The proposed framework can generate topic-focused crossword puzzles even with a small number of topic-derived words.
There could be educational applications, for example, choosing a textbook as a topic.

\paragraph{Improved Keyword Detection}

The quality of NER strongly propagates to crossword puzzles.
It is important to determine whether a noun is a suitable word for an answer.
A related research area is vocabulary acquisition support systems~\cite{Nation2006-bd,Ehara2010-au,Choffin2019-ea}.
Although this area has been studied specifically for second language learners, it is still relevant to our study in the sense that we encourage people to acquire new vocabulary words through crossword puzzles.

\paragraph{Improved Clue Generation}

High-quality clue generation is another important point.
Fill-in-the-blank questions are easy to create, but there is a challenge in controlling the level of difficulty~\cite{Kurdi2020-rz}.
For example, there is no guarantee that the answer is the only one that can be identified.
We can generate more reliable clues by utilizing question-answer datasets.
A rapidly evolving text generation approach, using pre-trained language models, is expected to improve clue quality~\cite{Du2017-hr}.
However, we must be careful of output that is not faithful to the facts.
We are exploring measures such as improving the quality of fine-tuning data, applying filters, and re-ranking output.
In addition, acquiring a text style specific to crossword puzzle clues is expected to enhance the quality of the output~\cite{Jin2022-go}.
We are attempting to characterize the text style of clues by referring to real-world crossword puzzles and interviewing their creators.

\paragraph{Improved Puzzle Combination Exploration}

The current prototype assumes the placement of black cells.
By considering optimization with variable black cell placements~\cite{Bulitko2021-jb,Dakowski2022-if}, we can generate crossword puzzles with more news-derived words.

\paragraph{Further User Testing}

Although we have performed a qualitative evaluation, larger-scale user testing would be desirable.
Examples include providing users with automatically generated crossword puzzles and investigating their interest in the news and its impact on education.

\section{Conclusion}
\label{sec:Conclusion}

This paper focused on the case of encouraging people's interest in news and proposed a framework for automatically generating news-centric crossword puzzles.
One of the contributions is that the educational objective of including more news-derived words is achieved as a constraint satisfaction and optimization problem.
Our experiments showed that news-centric crossword puzzles can be generated even with a small number of news-derived words.
We also demonstrated the prototype to approximately 20 people and described current issues and future directions.
We hope this paper accelerates research and practice in crossword puzzles for educational purposes.

\bibliographystyle{ACM-Reference-Format}

\begin{thebibliography}{33}


\ifx \showCODEN    \undefined \def \showCODEN     #1{\unskip}     \fi
\ifx \showDOI      \undefined \def \showDOI       #1{#1}\fi
\ifx \showISBNx    \undefined \def \showISBNx     #1{\unskip}     \fi
\ifx \showISBNxiii \undefined \def \showISBNxiii  #1{\unskip}     \fi
\ifx \showISSN     \undefined \def \showISSN      #1{\unskip}     \fi
\ifx \showLCCN     \undefined \def \showLCCN      #1{\unskip}     \fi
\ifx \shownote     \undefined \def \shownote      #1{#1}          \fi
\ifx \showarticletitle \undefined \def \showarticletitle #1{#1}   \fi
\ifx \showURL      \undefined \def \showURL       {\relax}        \fi
\providecommand\bibfield[2]{#2}
\providecommand\bibinfo[2]{#2}
\providecommand\natexlab[1]{#1}
\providecommand\showeprint[2][]{arXiv:#2}

\bibitem[Audemard et~al\mbox{.}(2020)]%
        {Audemard2020-yj}
\bibfield{author}{\bibinfo{person}{Gilles Audemard},
  \bibinfo{person}{Christophe Lecoutre}, {and} \bibinfo{person}{Mehdi Maamar}.}
  \bibinfo{year}{2020}\natexlab{}.
\newblock \showarticletitle{Segmented tables: An efficient modeling tool for
  constraint reasoning}. In \bibinfo{booktitle}{\emph{Proceedings of the 24th
  European Conference on Artificial Intelligence ({ECAI} 2020)}}.
  \bibinfo{publisher}{IOS Press}, \bibinfo{address}{Amsterdam, NY},
  \bibinfo{pages}{315--322}.
\newblock


\bibitem[Bonomo et~al\mbox{.}(2015)]%
        {Bonomo2015-ou}
\bibfield{author}{\bibinfo{person}{Douglas Bonomo}, \bibinfo{person}{Adrian~P
  Lauf}, {and} \bibinfo{person}{Roman Yampolskiy}.}
  \bibinfo{year}{2015}\natexlab{}.
\newblock \showarticletitle{A crossword puzzle generator using genetic
  algorithms with Wisdom of Artificial Crowds}. In
  \bibinfo{booktitle}{\emph{2015 Computer Games: {AI}, Animation, Mobile,
  Multimedia, Educational and Serious Games ({CGAMES})}}.
  \bibinfo{publisher}{IEEE}, \bibinfo{address}{Louisville, KY, USA},
  \bibinfo{pages}{44--49}.
\newblock


\bibitem[Botea(2007)]%
        {Botea2007-bn}
\bibfield{author}{\bibinfo{person}{Adi Botea}.}
  \bibinfo{year}{2007}\natexlab{}.
\newblock \showarticletitle{Crossword Grid Composition with A Hierarchical
  {CSP} Encoding}. In \bibinfo{booktitle}{\emph{Proceeding of the 6th
  International Workshop on Constraint Modelling and Reformulation
  (ModRef-07)}}. \bibinfo{address}{Providence, Rhode Island, USA},
  \bibinfo{numpages}{14}~pages.
\newblock


\bibitem[Botea and Bulitko(2021)]%
        {Botea2021-rb}
\bibfield{author}{\bibinfo{person}{Adi Botea} {and} \bibinfo{person}{Vadim
  Bulitko}.} \bibinfo{year}{2021}\natexlab{}.
\newblock \showarticletitle{Scaling Up Search with Partial Initial States in
  Optimization Crosswords}.
\newblock \bibinfo{journal}{\emph{The 14th International Symposium on
  Combinatorial Search}} \bibinfo{volume}{12}, \bibinfo{number}{1}
  (\bibinfo{date}{July} \bibinfo{year}{2021}), \bibinfo{pages}{20--27}.
\newblock


\bibitem[Brailsford et~al\mbox{.}(1999)]%
        {Brailsford1999-nw}
\bibfield{author}{\bibinfo{person}{Sally~C Brailsford},
  \bibinfo{person}{Chris~N Potts}, {and} \bibinfo{person}{Barbara~M Smith}.}
  \bibinfo{year}{1999}\natexlab{}.
\newblock \showarticletitle{Constraint satisfaction problems: Algorithms and
  applications}.
\newblock \bibinfo{journal}{\emph{European journal of operational research}}
  \bibinfo{volume}{119}, \bibinfo{number}{3} (\bibinfo{date}{Dec.}
  \bibinfo{year}{1999}), \bibinfo{pages}{557--581}.
\newblock


\bibitem[Bulitko and Botea(2021)]%
        {Bulitko2021-jb}
\bibfield{author}{\bibinfo{person}{Vadim Bulitko} {and} \bibinfo{person}{Adi
  Botea}.} \bibinfo{year}{2021}\natexlab{}.
\newblock \showarticletitle{Evolving Romanian Crossword Puzzles with Deep
  Learning and Heuristic Search}. In \bibinfo{booktitle}{\emph{Proceedings of
  the 2021 {IEEE} Conference on Games ({CoG})}}. \bibinfo{publisher}{IEEE},
  \bibinfo{address}{Copenhagen, Denmark}, \bibinfo{pages}{1--5}.
\newblock


\bibitem[Chali and Hasan(2015)]%
        {Chali2015-tm}
\bibfield{author}{\bibinfo{person}{Yllias Chali} {and} \bibinfo{person}{Sadid~A
  Hasan}.} \bibinfo{year}{2015}\natexlab{}.
\newblock \showarticletitle{Towards {Topic-to-Question} Generation}.
\newblock \bibinfo{journal}{\emph{Computational Linguistics}}
  \bibinfo{volume}{41}, \bibinfo{number}{1} (\bibinfo{date}{March}
  \bibinfo{year}{2015}), \bibinfo{pages}{1--20}.
\newblock


\bibitem[Choffin et~al\mbox{.}(2019)]%
        {Choffin2019-ea}
\bibfield{author}{\bibinfo{person}{B Choffin}, \bibinfo{person}{F Popineau},
  \bibinfo{person}{Y Bourda}, {and} \bibinfo{person}{J~J Vie}.}
  \bibinfo{year}{2019}\natexlab{}.
\newblock \showarticletitle{{DAS3H}: modeling student learning and forgetting
  for optimally scheduling distributed practice of skills}. In
  \bibinfo{booktitle}{\emph{Proceedings of the 12th International Conference on
  Educational Data Mining (EDM 2019)}}. \bibinfo{address}{Montréal, Canada},
  \bibinfo{pages}{29--38}.
\newblock


\bibitem[Claes and Quintelier(2009)]%
        {Claes2009-uk}
\bibfield{author}{\bibinfo{person}{Ellen Claes} {and} \bibinfo{person}{Ellen
  Quintelier}.} \bibinfo{year}{2009}\natexlab{}.
\newblock \showarticletitle{Newspapers in Education: a critical inquiry into
  the effects of using newspapers as teaching agents}.
\newblock \bibinfo{journal}{\emph{Educational Research}} \bibinfo{volume}{51},
  \bibinfo{number}{3} (\bibinfo{date}{Sept.} \bibinfo{year}{2009}),
  \bibinfo{pages}{341--363}.
\newblock


\bibitem[Dakowski et~al\mbox{.}(2022)]%
        {Dakowski2022-if}
\bibfield{author}{\bibinfo{person}{Jakub Dakowski}, \bibinfo{person}{Piotr
  Jaworski}, {and} \bibinfo{person}{Waldemar Wojna}.}
  \bibinfo{year}{2022}\natexlab{}.
\newblock \showarticletitle{Quick generation of crosswords using
  concatenation}. In \bibinfo{booktitle}{\emph{Proceedings of the 2022 {IEEE}
  Conference on Games ({CoG})}}. \bibinfo{publisher}{IEEE},
  \bibinfo{address}{Beijing, China}, \bibinfo{pages}{590--593}.
\newblock


\bibitem[De~Kegel and Haahr(2020)]%
        {De_Kegel2020-jl}
\bibfield{author}{\bibinfo{person}{Barbara De~Kegel} {and}
  \bibinfo{person}{Mads Haahr}.} \bibinfo{year}{2020}\natexlab{}.
\newblock \showarticletitle{Procedural Puzzle Generation: A Survey}.
\newblock \bibinfo{journal}{\emph{IEEE Transactions on Computational
  Intelligence in AI and Games}} \bibinfo{volume}{12}, \bibinfo{number}{1}
  (\bibinfo{date}{March} \bibinfo{year}{2020}), \bibinfo{pages}{21--40}.
\newblock


\bibitem[DeRoche(1981)]%
        {DeRoche1981-um}
\bibfield{author}{\bibinfo{person}{Edward~F DeRoche}.}
  \bibinfo{year}{1981}\natexlab{}.
\newblock \showarticletitle{Newspapers in Education: What We Know}.
\newblock \bibinfo{journal}{\emph{Newsp. Res. J.}} \bibinfo{volume}{2},
  \bibinfo{number}{3} (\bibinfo{date}{April} \bibinfo{year}{1981}),
  \bibinfo{pages}{59--63}.
\newblock


\bibitem[Du et~al\mbox{.}(2017)]%
        {Du2017-hr}
\bibfield{author}{\bibinfo{person}{Xinya Du}, \bibinfo{person}{Junru Shao},
  {and} \bibinfo{person}{Claire Cardie}.} \bibinfo{year}{2017}\natexlab{}.
\newblock \showarticletitle{Learning to Ask: Neural Question Generation for
  Reading Comprehension}. In \bibinfo{booktitle}{\emph{Proceedings of the 55th
  Annual Meeting of the Association for Computational Linguistics (Volume 1:
  Long Papers)}}. \bibinfo{publisher}{Association for Computational
  Linguistics}, \bibinfo{address}{Vancouver, Canada},
  \bibinfo{pages}{1342--1352}.
\newblock


\bibitem[Ehara et~al\mbox{.}(2010)]%
        {Ehara2010-au}
\bibfield{author}{\bibinfo{person}{Yo Ehara}, \bibinfo{person}{Nobuyuki
  Shimizu}, \bibinfo{person}{Takashi Ninomiya}, {and} \bibinfo{person}{Hiroshi
  Nakagawa}.} \bibinfo{year}{2010}\natexlab{}.
\newblock \showarticletitle{Personalized reading support for second-language
  web documents by collective intelligence}. In
  \bibinfo{booktitle}{\emph{Proceedings of the 15th international conference on
  Intelligent user interfaces}} (Hong Kong, China) \emph{(\bibinfo{series}{IUI
  '10})}. \bibinfo{publisher}{Association for Computing Machinery},
  \bibinfo{address}{New York, NY, USA}, \bibinfo{pages}{51--60}.
\newblock


\bibitem[Esteche et~al\mbox{.}(2016)]%
        {Esteche2016-db}
\bibfield{author}{\bibinfo{person}{Jennifer Esteche}, \bibinfo{person}{Romina
  Romero}, \bibinfo{person}{Luis Chiruzzo}, {and} \bibinfo{person}{Aiala
  Ros{\'a}}.} \bibinfo{year}{2016}\natexlab{}.
\newblock \showarticletitle{Automatic definition extraction and crosswords
  generation from news text}. In \bibinfo{booktitle}{\emph{Proceedings of the
  2016 {XLII} Latin American Computing Conference ({CLEI})}}.
  \bibinfo{publisher}{IEEE}, \bibinfo{address}{Valparaíso, Chile},
  \bibinfo{pages}{1--8}.
\newblock


\bibitem[Ferrer-Conill et~al\mbox{.}(2020)]%
        {Ferrer-Conill2020-ic}
\bibfield{author}{\bibinfo{person}{Raul Ferrer-Conill},
  \bibinfo{person}{Maxwell Foxman}, \bibinfo{person}{Janet Jones},
  \bibinfo{person}{Tanja Sihvonen}, {and} \bibinfo{person}{Marko Siitonen}.}
  \bibinfo{year}{2020}\natexlab{}.
\newblock \showarticletitle{Playful approaches to news engagement}.
\newblock \bibinfo{journal}{\emph{Convergence}} \bibinfo{volume}{26},
  \bibinfo{number}{3} (\bibinfo{date}{June} \bibinfo{year}{2020}),
  \bibinfo{pages}{457--469}.
\newblock


\bibitem[Franklin et~al\mbox{.}(2003)]%
        {Franklin2003-cl}
\bibfield{author}{\bibinfo{person}{Sue Franklin}, \bibinfo{person}{Mary Peat},
  {and} \bibinfo{person}{Alison Lewis}.} \bibinfo{year}{2003}\natexlab{}.
\newblock \showarticletitle{Non-traditional interventions to stimulate
  discussion: the use of games and puzzles}.
\newblock \bibinfo{journal}{\emph{Journal of biological education}}
  \bibinfo{volume}{37}, \bibinfo{number}{2} (\bibinfo{date}{March}
  \bibinfo{year}{2003}), \bibinfo{pages}{79--84}.
\newblock


\bibitem[Ginsberg et~al\mbox{.}(1990)]%
        {Ginsberg1990-em}
\bibfield{author}{\bibinfo{person}{Matthew~L Ginsberg},
  \bibinfo{person}{Michael Frank}, \bibinfo{person}{Michael~P Halpin}, {and}
  \bibinfo{person}{Mark~C Torrance}.} \bibinfo{year}{1990}\natexlab{}.
\newblock \showarticletitle{Search lessons learned from crossword puzzles}. In
  \bibinfo{booktitle}{\emph{Proceedings of the 8th National conference on
  Artificial intelligence - Volume 1}} (Boston, Massachusetts)
  \emph{(\bibinfo{series}{AAAI'90})}. \bibinfo{publisher}{AAAI Press},
  \bibinfo{address}{Boston, Massachusett, USA}, \bibinfo{pages}{210--215}.
\newblock


\bibitem[Heilman and Smith(2010)]%
        {Heilman2010-ie}
\bibfield{author}{\bibinfo{person}{Michael Heilman} {and}
  \bibinfo{person}{Noah~A Smith}.} \bibinfo{year}{2010}\natexlab{}.
\newblock \showarticletitle{Good Question! Statistical Ranking for Question
  Generation}. In \bibinfo{booktitle}{\emph{Human Language Technologies: The
  2010 Annual Conference of the North {A}merican Chapter of the Association for
  Computational Linguistics}}. \bibinfo{publisher}{Association for
  Computational Linguistics}, \bibinfo{address}{Los Angeles, California},
  \bibinfo{pages}{609--617}.
\newblock


\bibitem[Jin et~al\mbox{.}(2022)]%
        {Jin2022-go}
\bibfield{author}{\bibinfo{person}{Di Jin}, \bibinfo{person}{Zhijing Jin},
  \bibinfo{person}{Zhiting Hu}, \bibinfo{person}{Olga Vechtomova}, {and}
  \bibinfo{person}{Rada Mihalcea}.} \bibinfo{year}{2022}\natexlab{}.
\newblock \showarticletitle{Deep Learning for Text Style Transfer: A Survey}.
\newblock \bibinfo{journal}{\emph{Computational Linguistics}}
  \bibinfo{volume}{48}, \bibinfo{number}{1} (\bibinfo{date}{March}
  \bibinfo{year}{2022}), \bibinfo{pages}{155--205}.
\newblock


\bibitem[Khaira(2021)]%
        {Khaira2021-xe}
\bibfield{author}{\bibinfo{person}{Mutia Khaira}.}
  \bibinfo{year}{2021}\natexlab{}.
\newblock \showarticletitle{The Effectiveness of Crossword Puzzle in Improving
  Mufradat Skills}.
\newblock \bibinfo{journal}{\emph{Tanwir Arabiyyah: Arabic As Foreign Language
  Journal}} \bibinfo{volume}{1}, \bibinfo{number}{2} (\bibinfo{date}{Dec.}
  \bibinfo{year}{2021}), \bibinfo{pages}{55--62}.
\newblock


\bibitem[Kurdi et~al\mbox{.}(2020)]%
        {Kurdi2020-rz}
\bibfield{author}{\bibinfo{person}{Ghader Kurdi}, \bibinfo{person}{Jared Leo},
  \bibinfo{person}{Bijan Parsia}, \bibinfo{person}{Uli Sattler}, {and}
  \bibinfo{person}{Salam Al-Emari}.} \bibinfo{year}{2020}\natexlab{}.
\newblock \showarticletitle{A Systematic Review of Automatic Question
  Generation for Educational Purposes}.
\newblock \bibinfo{journal}{\emph{International Journal of Artificial
  Intelligence in Education}} \bibinfo{volume}{30}, \bibinfo{number}{1}
  (\bibinfo{date}{March} \bibinfo{year}{2020}), \bibinfo{pages}{121--204}.
\newblock


\bibitem[Littman et~al\mbox{.}(2002)]%
        {Littman2002-sm}
\bibfield{author}{\bibinfo{person}{Michael~L Littman}, \bibinfo{person}{Greg~A
  Keim}, {and} \bibinfo{person}{Noam Shazeer}.}
  \bibinfo{year}{2002}\natexlab{}.
\newblock \showarticletitle{A probabilistic approach to solving crossword
  puzzles}.
\newblock \bibinfo{journal}{\emph{Artificial intelligence}}
  \bibinfo{volume}{134}, \bibinfo{number}{1-2} (\bibinfo{date}{Jan.}
  \bibinfo{year}{2002}), \bibinfo{pages}{23--55}.
\newblock


\bibitem[Mazlack(1976)]%
        {Mazlack1976-wz}
\bibfield{author}{\bibinfo{person}{Lawrence~J Mazlack}.}
  \bibinfo{year}{1976}\natexlab{}.
\newblock \showarticletitle{Computer construction of crossword puzzles using
  precedence relationships}.
\newblock \bibinfo{journal}{\emph{Artificial intelligence}}
  \bibinfo{volume}{7}, \bibinfo{number}{1} (\bibinfo{date}{March}
  \bibinfo{year}{1976}), \bibinfo{pages}{1--19}.
\newblock


\bibitem[Meehan and Gray(1997)]%
        {meehan1997constructing}
\bibfield{author}{\bibinfo{person}{Gary Meehan} {and} \bibinfo{person}{Peter
  Gray}.} \bibinfo{year}{1997}\natexlab{}.
\newblock \showarticletitle{Constructing crossword grids: Use of heuristics vs
  constraints}.
\newblock \bibinfo{journal}{\emph{Proceedings of Expert Systems}}
  \bibinfo{volume}{97} (\bibinfo{year}{1997}), \bibinfo{pages}{159--174}.
\newblock


\bibitem[Nation(2006)]%
        {Nation2006-bd}
\bibfield{author}{\bibinfo{person}{I Nation}.} \bibinfo{year}{2006}\natexlab{}.
\newblock \showarticletitle{How large a vocabulary is needed for reading and
  listening?}
\newblock \bibinfo{journal}{\emph{La revue canadienne des langues vivantes [The
  Canadian modern language review]}} \bibinfo{volume}{63}, \bibinfo{number}{1}
  (\bibinfo{date}{Sept.} \bibinfo{year}{2006}), \bibinfo{pages}{59--82}.
\newblock


\bibitem[Ritonga et~al\mbox{.}(2021)]%
        {Ritonga2021-wq}
\bibfield{author}{\bibinfo{person}{Apri~Wardana Ritonga},
  \bibinfo{person}{Mahyudin Ritonga}, \bibinfo{person}{Vini~Wela Septiana},
  {and} \bibinfo{person}{{Mahmud}}.} \bibinfo{year}{2021}\natexlab{}.
\newblock \showarticletitle{Crossword puzzle as a learning media during the
  covid-19 pandemic: {HOTS}, {MOTS} or {LOTS}?}
\newblock \bibinfo{journal}{\emph{Journal of physics. Conference series}}
  \bibinfo{volume}{1933}, \bibinfo{number}{1} (\bibinfo{date}{June}
  \bibinfo{year}{2021}), \bibinfo{pages}{012126}.
\newblock


\bibitem[Sun et~al\mbox{.}(2018)]%
        {Sun2018-pm}
\bibfield{author}{\bibinfo{person}{Xingwu Sun}, \bibinfo{person}{Jing Liu},
  \bibinfo{person}{Yajuan Lyu}, \bibinfo{person}{Wei He},
  \bibinfo{person}{Yanjun Ma}, {and} \bibinfo{person}{Shi Wang}.}
  \bibinfo{year}{2018}\natexlab{}.
\newblock \showarticletitle{Answer-focused and Position-aware Neural Question
  Generation}. In \bibinfo{booktitle}{\emph{Proceedings of the 2018 Conference
  on Empirical Methods in Natural Language Processing}}.
  \bibinfo{publisher}{Association for Computational Linguistics},
  \bibinfo{address}{Brussels, Belgium}, \bibinfo{pages}{3930--3939}.
\newblock


\bibitem[Suryawati et~al\mbox{.}(2018)]%
        {Suryawati2018-cv}
\bibfield{author}{\bibinfo{person}{Evi Suryawati}, \bibinfo{person}{Fitra
  Suzanti}, \bibinfo{person}{Suwondo Suwondo}, {and} \bibinfo{person}{Yustina
  Yustina}.} \bibinfo{year}{2018}\natexlab{}.
\newblock \showarticletitle{The implementation of school-literacy-movement:
  Integrating scientific literacy, characters, and {HOTS} in science learning}.
\newblock \bibinfo{journal}{\emph{JPBI (Jurnal Pendidikan Biologi Indonesia)}}
  \bibinfo{volume}{4}, \bibinfo{number}{3} (\bibinfo{date}{Nov.}
  \bibinfo{year}{2018}), \bibinfo{pages}{215--224}.
\newblock


\bibitem[Tarjan(1971)]%
        {Tarjan1971-hx}
\bibfield{author}{\bibinfo{person}{Robert Tarjan}.}
  \bibinfo{year}{1971}\natexlab{}.
\newblock \showarticletitle{Depth-first search and linear graph algorithms}. In
  \bibinfo{booktitle}{\emph{Proceedings of the 12th Annual Symposium on
  Switching and Automata Theory (swat 1971)}}. \bibinfo{publisher}{IEEE},
  \bibinfo{address}{East Lansing, MI, USA}, \bibinfo{pages}{114--121}.
\newblock


\bibitem[Wang et~al\mbox{.}(2022)]%
        {Wang2022-qa}
\bibfield{author}{\bibinfo{person}{Xu Wang}, \bibinfo{person}{Simin Fan},
  \bibinfo{person}{Jessica Houghton}, {and} \bibinfo{person}{Lu Wang}.}
  \bibinfo{year}{2022}\natexlab{}.
\newblock \showarticletitle{Towards {Process-Oriented}, Modular, and Versatile
  Question Generation that Meets Educational Needs}. In
  \bibinfo{booktitle}{\emph{Proceedings of the 2022 Conference of the North
  American Chapter of the Association for Computational Linguistics: Human
  Language Technologies}}. \bibinfo{publisher}{Association for Computational
  Linguistics}, \bibinfo{address}{Seattle, United States},
  \bibinfo{pages}{291--302}.
\newblock


\bibitem[Yadav and Bethard(2018)]%
        {Yadav2018-gb}
\bibfield{author}{\bibinfo{person}{Vikas Yadav} {and} \bibinfo{person}{Steven
  Bethard}.} \bibinfo{year}{2018}\natexlab{}.
\newblock \showarticletitle{A Survey on Recent Advances in Named Entity
  Recognition from Deep Learning models}. In
  \bibinfo{booktitle}{\emph{Proceedings of the 27th International Conference on
  Computational Linguistics}}. \bibinfo{publisher}{Association for
  Computational Linguistics}, \bibinfo{address}{Santa Fe, New Mexico, USA},
  \bibinfo{pages}{2145--2158}.
\newblock


\bibitem[Zhang and Bansal(2019)]%
        {Zhang2019-nz}
\bibfield{author}{\bibinfo{person}{Shiyue Zhang} {and} \bibinfo{person}{Mohit
  Bansal}.} \bibinfo{year}{2019}\natexlab{}.
\newblock \showarticletitle{Addressing Semantic Drift in Question Generation
  for {Semi-Supervised} Question Answering}. In
  \bibinfo{booktitle}{\emph{Proceedings of the 2019 Conference on Empirical
  Methods in Natural Language Processing and the 9th International Joint
  Conference on Natural Language Processing ({EMNLP-IJCNLP})}}.
  \bibinfo{publisher}{Association for Computational Linguistics},
  \bibinfo{address}{Hong Kong, China}, \bibinfo{pages}{2495--2509}.
\newblock


\end{thebibliography}

\end{document}